\documentclass[letterpaper, 10 pt, conference]{ieeeconf}
\IEEEoverridecommandlockouts
\usepackage{times}
\usepackage{graphics}
\usepackage{graphicx}
\usepackage{subfigure}
\usepackage{amsmath,amssymb,amsopn,amstext,amsfonts}
\usepackage{cancel}
\usepackage[space]{cite}
\usepackage{pdfsync}
\usepackage{balance}
\usepackage{color}
\usepackage{mathtools}
\usepackage{bm}

\usepackage{diagbox}
\usepackage{float}
\usepackage{epstopdf}
\usepackage{pifont}
\usepackage{fixltx2e}
\usepackage{amsmath}
\usepackage{multirow}
\usepackage{url}
\usepackage{verbatim}
\usepackage{caption}
\usepackage{adjustbox}
\usepackage{booktabs}
\usepackage{threeparttable}
\usepackage{makecell}
\usepackage{tabularx}
\usepackage{arydshln}
\usepackage[normalem]{ulem}
\usepackage{algorithm}
\usepackage{ifthen}
\usepackage{algorithmic}
\usepackage{array}
\usepackage{textcomp}
\usepackage{stfloats}

\newcommand{\etal}{\textit{et al}.}
\newcommand{\ie}{\textit{i}.\textit{e}.}
\newcommand{\eg}{\textit{e}.\textit{g}.}

\makeatletter
\let\NAT@parse\undefined
\makeatother
\usepackage[linkcolor=red,citecolor=green,urlcolor=red,colorlinks=true]{hyperref}

\title{\LARGE \bf IMU-Aided Event-based Stereo Visual Odometry}
\author{Junkai Niu$^{\ast}$, Sheng Zhong$^{\ast}$, Yi Zhou
\thanks{All authors are with the Neuromorphic Automation and Intelligence Lab (NAIL) at School of Robotics, Hunan University, Changsha, China. }
\thanks{$\ast$ denotes equal contribution.}
\thanks{Corresponding author: Yi Zhou. Email: {\tt\small eeyzhou@hnu.edu.cn}.}
\thanks{This work was supported by the National Key Research and Development Project of China under Grant 2023YFB4706600.
}
}

\global\long\def\bc{\mathbf{c}}

\begin{document}
\maketitle
\thispagestyle{empty}
\pagestyle{empty}

\begin{abstract}
\label{sec:abstract}
Direct methods for event-based visual odometry solve the mapping and camera pose tracking sub-problems by establishing implicit data association in a way that the generative model of events is exploited.
The main bottlenecks faced by state-of-the-art work in this field include the high computational complexity of mapping and the limited accuracy of tracking.
In this paper, we improve our previous direct pipeline \textit{Event-based Stereo Visual Odometry} in terms of accuracy and efficiency.
To speed up the mapping operation, we propose an efficient strategy of edge-pixel sampling according to the local dynamics of events.
The mapping performance in terms of completeness and local smoothness is also improved by combining the temporal stereo results and the static stereo results.
To circumvent the degeneracy issue of camera pose tracking in recovering the yaw component of general 6-DoF motion, we introduce as a prior the gyroscope measurements via pre-integration.
Experiments on publicly available datasets justify our improvement.
We release our pipeline as an open-source software for future research in this field.
\end{abstract}

\section*{Multimedia Material}

\noindent Supplemental Video: {\small\url{https://youtu.be/hy25-nExD0E}}\\
\noindent Code: {\small\url{https://github.com/NAIL-HNU/ESVIO_AA.git}}

\section{Introduction}
\label{sec: introduction}

Neuromorphic event-based cameras are bio-inspired visual sensors with asynchronous pixels that report only intensity changes (called ``events'').
Endowed with microsecond temporal resolution and up to 160 dB dynamic range \cite{Lichtsteiner08ssc}, event cameras are qualified to deal with challenging scenarios that are inaccessible to standard cameras, such as high-speed and/or high-dynamic-range (HDR) tracking~\cite{Mueggler14iros,Lagorce15tnnls,Zhu17icra,Gallego17pami,Gallego17ral,Mueggler17tro,Gehrig18eccv}, control~\cite{Conradt09iscas,Delbruck13fns} and Simultaneous Localization and Mapping (SLAM)~\cite{Rebecq16bmvc,Kim16eccv,Rebecq17ijcv,Rebecq17ral,Rebecq17bmvc,Rosinol18ral,zhou2021event}.

Like its standard-vision counterparts, event-based visual odometry (VO) or SLAM also aims at solving simultaneously the mapping and tracking sub-problems in a recursive manner.
The main challenge therein is to extract and maintain effective data association in the event stream, from which the depth information and ego motion can be inferred.
From the perspective of how such data association is established, existing methods, including event-based visual inertial odometry, can be divided into two categories: feature-based methods and direct methods.

\textbf{Feature-based Methods:}
To build on top of existing feature-based VO/SLAM pipelines (\eg, \cite{klein2007parallel, MurArtal15tro}) using standard cameras, researchers have resorted to developing hand-crafted features from event data, such as event corners~\cite{Vasco16iros, Mueggler17bmvc,Alzugaray18ral,Li19iros}, which are typically adapted from the original Harris~\cite{Harris88} and FAST~\cite{Rosten06eccv} methods.
Additionally, strategies for tracking event corners are presented by \cite{Alzugaray18ral, Alzugaray18threedv}.
Such event features enable straightforward application of epipolar geometry~\cite{hartley1997defense, nister2004efficient} and Perspective-\textit{n}-Point (PnP) methods~\cite{li2012robust, kneip2014upnp}.
Although the success of these feature-based solutions (\eg,~\cite{hadviger2021feature}) has been witnessed to some extent, event features, however, are not as theoretically robust as their standard-vision counterparts.
This is due to the camera-velocity-dependent nature of event data, which sometimes leads to incomplete observation of junctions.
Consequently, feature matching can easily fail in a sudden variation of the event camera's velocity.
Another mainstream strategy for feature detection and tracking is inspired by the motion compensation method \cite{Gallego18cvpr}, a unified pipeline for event-based geometric model fitting.
This strategy is widely witnessed in event-based VIO pipelines~\cite{Zhu17cvpr, Rebecq17bmvc}, which typically build features from motion-compensated event sets~\cite{Zhu17icra} or event images~\cite{Gallego18cvpr}, and furthermore, fuse with inertial measurements by means of either the Kalman filter~\cite{Mourikis07icra} or the keyframe-based nonlinear optimization~\cite{leutenegger2013keyframe}.

\begin{figure}[t]
  \centering
  \includegraphics[width=0.98\linewidth]{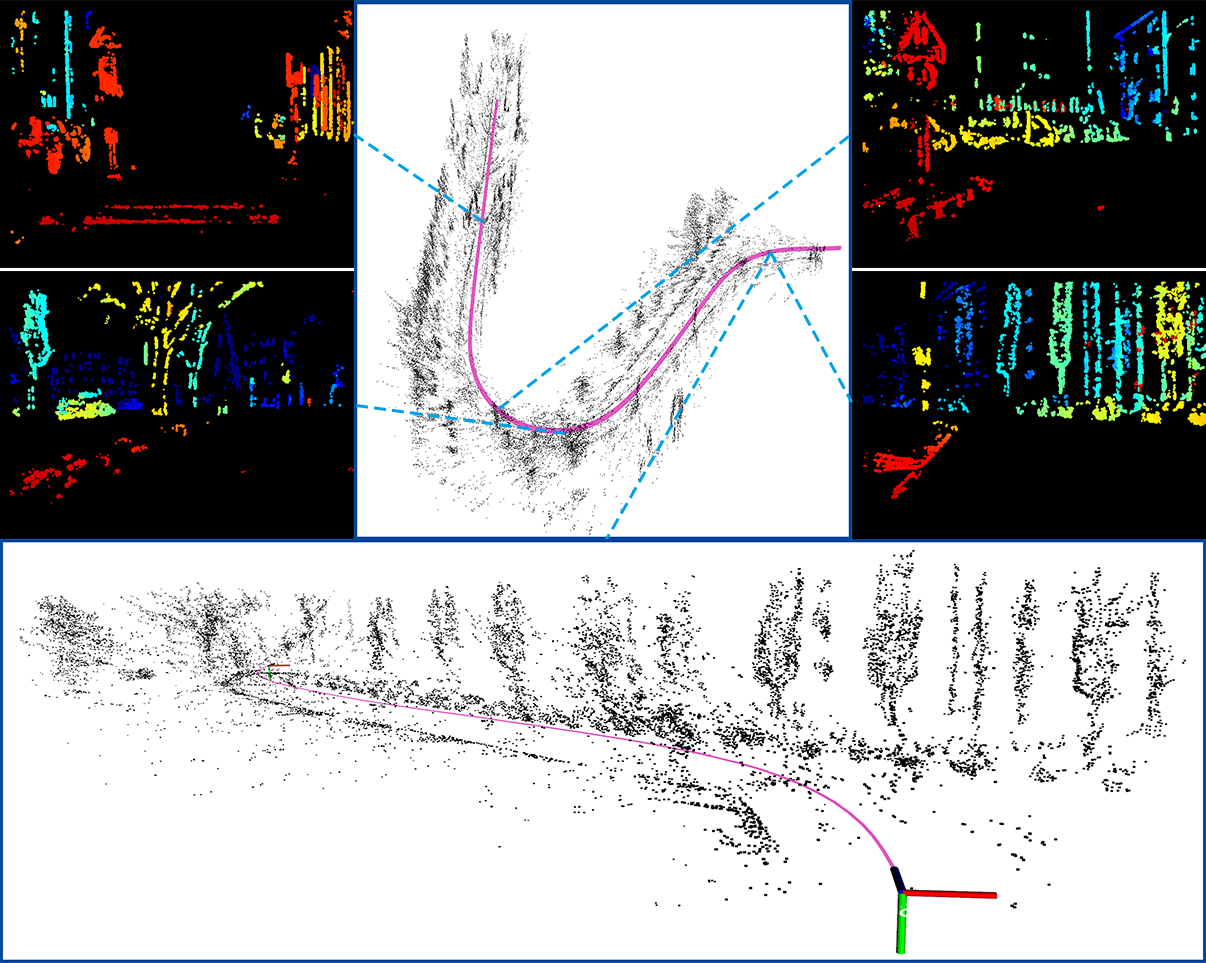}
   \captionof{figure}{Illustration of running the proposed system on \textit{DSEC}~\cite{Gehrig21ral} dataset (seq. \textit{dsec\_city04\_a}). 
   Middle and bottom: The reconstructed point cloud and resulting trajectory seen from different view points.
   Left and right: Recovered inverse depth maps at reference perspectives.}
\label{fig:eye catcher}
\vspace{-1.5em}
\end{figure}

\textbf{Direct Methods:}
Unlike feature-based methods, direct methods refer to those that implicitly establish data associations by exploiting the generative model of events in some ways.
Based on the constant-brightness assumption in the log intensity domain, Kim \etal \cite{Kim16eccv} propose the first direct method that implements three interleaved probabilistic filters solving the sub-problems of mapping, camera pose tracking, and additionally, recovering the log intensity information.
To justify that recovering the intensity information is not mandatory, Rebecq \etal \cite{Rebecq17ral} propose a pure geometric method.
The proposed mapping approach determines the 3D location of structures by searching the maximum ray intersection
in the disparity space image (DSI), and the camera pose is estimated through a 3D-2D registration process, in which the 3D edge map is aligned to the synthesized event map.
These two pioneering frameworks are, however, limited by the requirements of gentle motion in the initialization and slow expansion of the local map.
Hence, neither of them has been justified on data collected using a mobile platform.
To overcome these issues, Zhou \etal~present the first event-based VO pipeline (ESVO) using a stereo event camera~\cite{zhou2021event}.
The method exploits spatio-temporal consistency of the events across the image planes of the cameras to solve both localization and mapping sub-problems of visual odometry.
Nevertheless, ESVO dose not achieve near real-time performance once the spatial resolution of event cameras is $640 \times 480$ pixels or larger.
This is mainly due to the large number of redundant operations in the mapping, which is originally caused by the way that edge pixels are determined.
Besides, we observe that ESVO's tracking method cannot fully recover the yaw component in general 6-DoF motion, 
which degrades the accuracy of the recovered trajectory.

The goal of this paper is to lift the above-mentioned limitations of the original ESVO framework.
We extend ESVO and present a direct method for visual-inertial odometry with a stereo event camera, , as illustrated in~\ref{fig:eye catcher}.
The proposed system achieves better performance of mapping and tracking than ESVO in terms of accuracy and efficiency, due to the following efforts.\\
\textbf{\textit{Contribution.}}
\begin{itemize}
    \item A novel image-like representation of events, called adaptive accumulation (AA) of events, which is used for efficient determination of pixel locations associated to instantaneous edges.
    \item An improved solution to the mapping sub-problem by leveraging both the temporal stereo and static stereo configurations;
    \item An IMU-aided solution to the camera pose tracking sub-problem that overcomes the insensitivity to the yaw component of general 6-DoF motion in the 3D-2D spatio-temporal registration. 
\end{itemize}

The remaining paper is organized as follows.
We first discuss our method by detailing each item listed in the contribution (Sec.~\ref{sec:method}).
Then the experimental evaluation is provided in Sec.~\ref{sec:evaluation}, and finally the conclusion is made in Sec.~\ref{sec:conclusion}.

\section{Methodology}
\label{sec:method}

We detail our method in this section.
First, we present a pre-processing method for event data that samples efficiently a number of edge-pixel candidates (Sec.~\ref{subsec:Adaptive Accumulation of Events}).
Second, we discuss how to recover the depth information of missing structures in the original ESVO pipeline and further achieve a more complete mapping result by merging results of temporal stereo and static stereo methods (Sec.~\ref{subsec:mapping}).
Third, we discuss how to improve the 3D-2D spatio-temporal registration by leveraging inertial measurements as a motion prior (Sec.~\ref{subsec:tracking}).
Finally, we overview the proposed system and discuss the implementation detail (Sec.~\ref{subsec:system}).

\subsection{Adaptive Accumulation of Events}
\label{subsec:Adaptive Accumulation of Events}

The computational efficiency of ESVO's mapping method is limited by several aspects.
One of them is the way that the pixel locations of instantaneous edges are determined.
In ESVO, the instantaneous edge map in a virtual reference frame is created by applying motion compensation to events occurred within a short time interval (\eg, 10 ms).
This operation will become too computationally expensive as a pre-processing step when the event streaming rate exceeds a certain range\footnote{The event streaming rate is in proportion to the scene dynamics, the scene texture and the spatial resolution of the event camera}.
Besides, we observe that the extracted events are concentrated in the regions of significant optical flow.
To alleviate such an uneven distribution of edge pixels, it is necessary to sample a large number of points, which is redundant and in turn becomes a computation burden to mapping.
Therefore, a more efficient way is needed to determine pixels of instantaneous edges.

We propose a novel method inspired by \cite{Rebecq17ral,liu2018adaptive}.
In \cite{Rebecq17ral}, the synthesized event map obtained by a naive accumulation\footnote{Naive accumulation of events refers to plainly visualizing events' spatial information (\ie~image coordinates) on the image plane without any transformation in the spatio-temporal domain.} of events can be used as an approximate edge map.
However, this approximation can become severely inaccurate (\ie, blurred edges or invisible edges) when there is a significant parallax in the scene.
This is because a global threshold for accumulation cannot handle different local dynamics of events.
To this end, we propose a method called adaptive accumulation (AA) that can control the amount of events to be accumulated according to the local event dynamics.
Intuitively, the more intensive the local event dynamics, the shorter the time interval for event accumulation.
Different from \cite{liu2018adaptive} which directly uses the number of events as the metric to control the time length for event accumulation, we apply the contrast of event image \cite{Gallego18cvpr}.
The contrast of image can be quantified by a variety of dispersion metrics, and we simply use the variance loss because of its advantageous accuracy and computation complexity over other alternatives~\cite{Gallego19cvpr}.
Although the contrast of event image \cite{Gallego18cvpr} monotonically increases as the number of accumulated events increases, we observe that, for a fixed-size local patch, the accumulation result by a certain contrast value ($\beta$) can deliver enough visual information before getting obviously blurred.

As shown in Algorithm~\ref{alg:AA algorithm}, an AA map is generated as follows.
First, we divide the image plane evenly to several small blocks (e.g., block size $w \times w$), and the accumulation of events is carried out in each block independently.
For each block area, the contrast of event image \cite{Gallego18cvpr} is evaluated at a fixed time interval $\delta t$, and the computation terminates if the contrast reaches a certain threshold $\beta$.
In this way, patches with different local event dynamics will possibly have a similar number of accumulated events.

\begin{figure}[t]
	\centering
    \subfigure[Time Surface~\cite{Delbruck08issle}]{\includegraphics[width=0.23\textwidth]{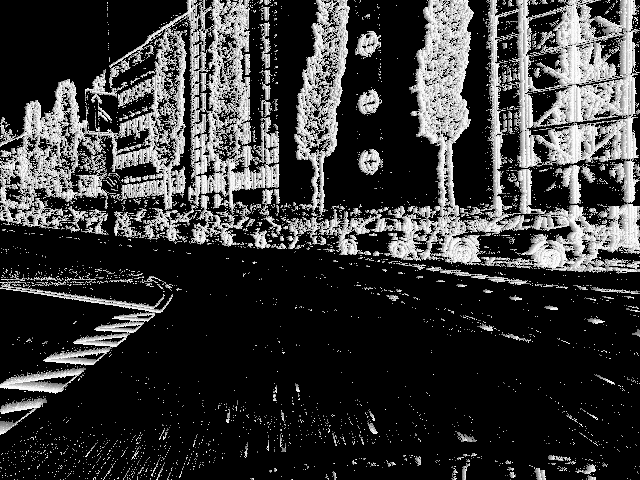}}\hspace{5pt}
	\subfigure[TOS~\cite{glover2021luvharris}]{\includegraphics[width=0.23\textwidth]{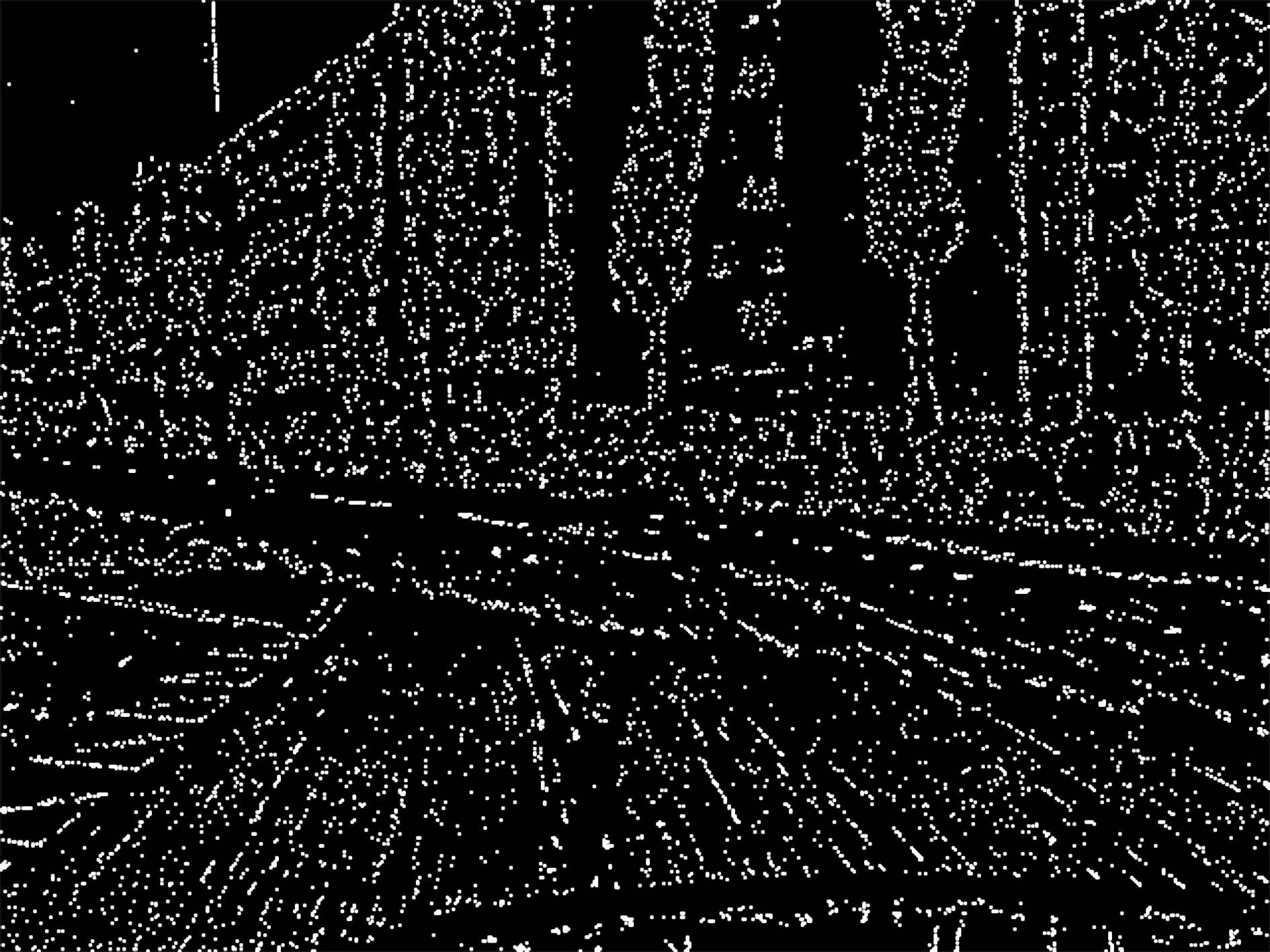}}\\
	\subfigure[SILC~\cite{Manderscheid19cvpr}]{\includegraphics[width=0.23\textwidth]{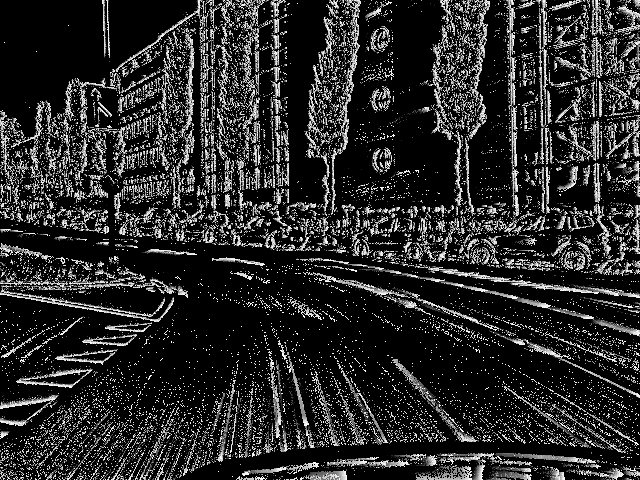}}\hspace{5pt}
	\subfigure[Adaptive Accumulation]{\includegraphics[width=0.23\textwidth]{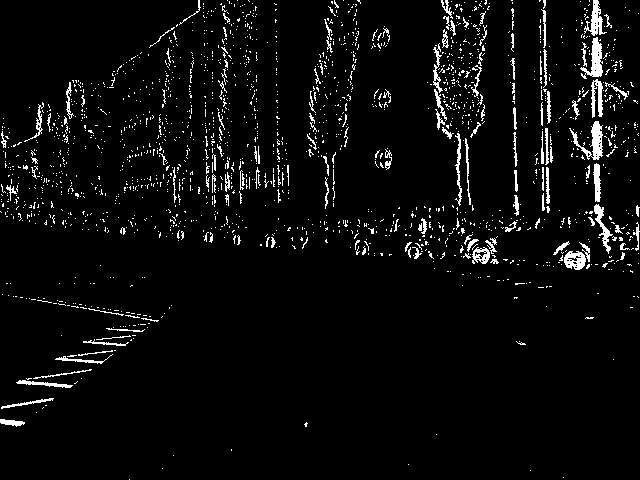}}
	\caption{\textit{Comparison of resulting edge maps from TS~\cite{Delbruck08issle}, TOS~\cite{glover2021luvharris}, SILC~\cite{Manderscheid19cvpr}, and AA, respectively.}} 
    \label{fig:image-like representation comparision}
\end{figure}

\begin{figure}[t]
	\centering
	\subfigure[ESVO (10k pts.)]{\includegraphics[width=0.15\textwidth]{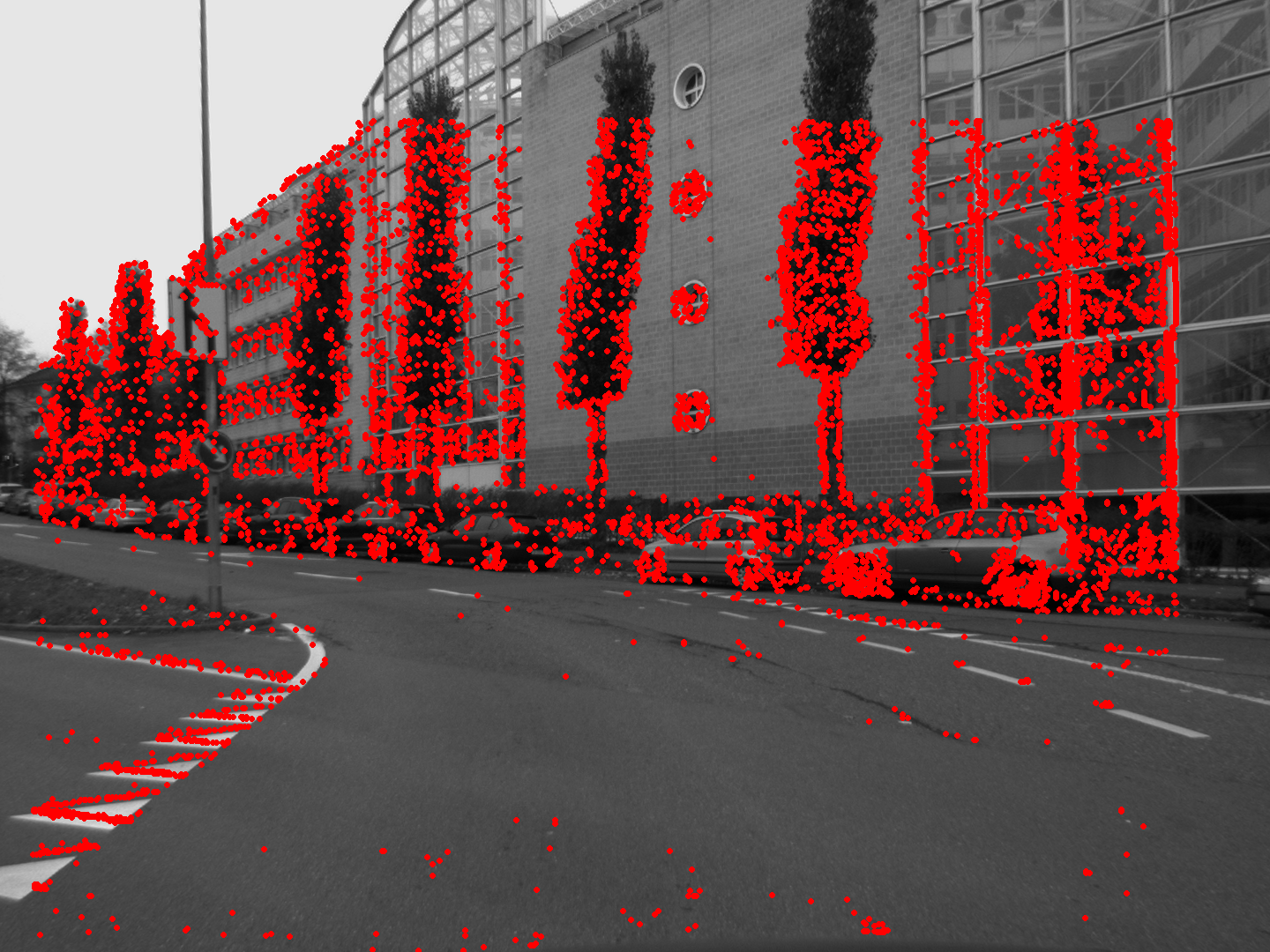}}\hspace{1pt}
    \subfigure[ESVO (5k pts.)]{\includegraphics[width=0.15\textwidth]{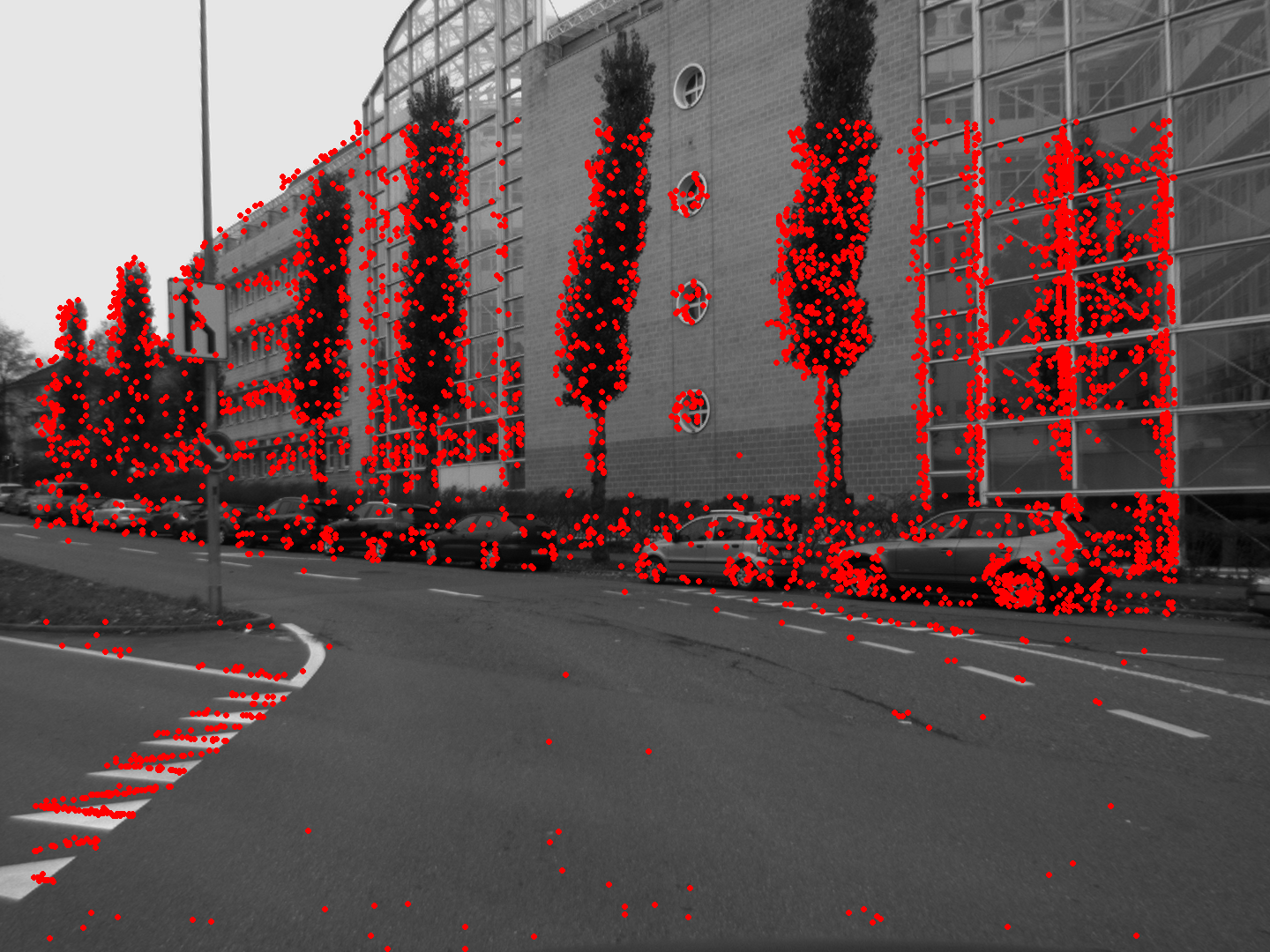}}\hspace{1pt}
	\subfigure[Ours (5k pts.)]{\includegraphics[width=0.15\textwidth]{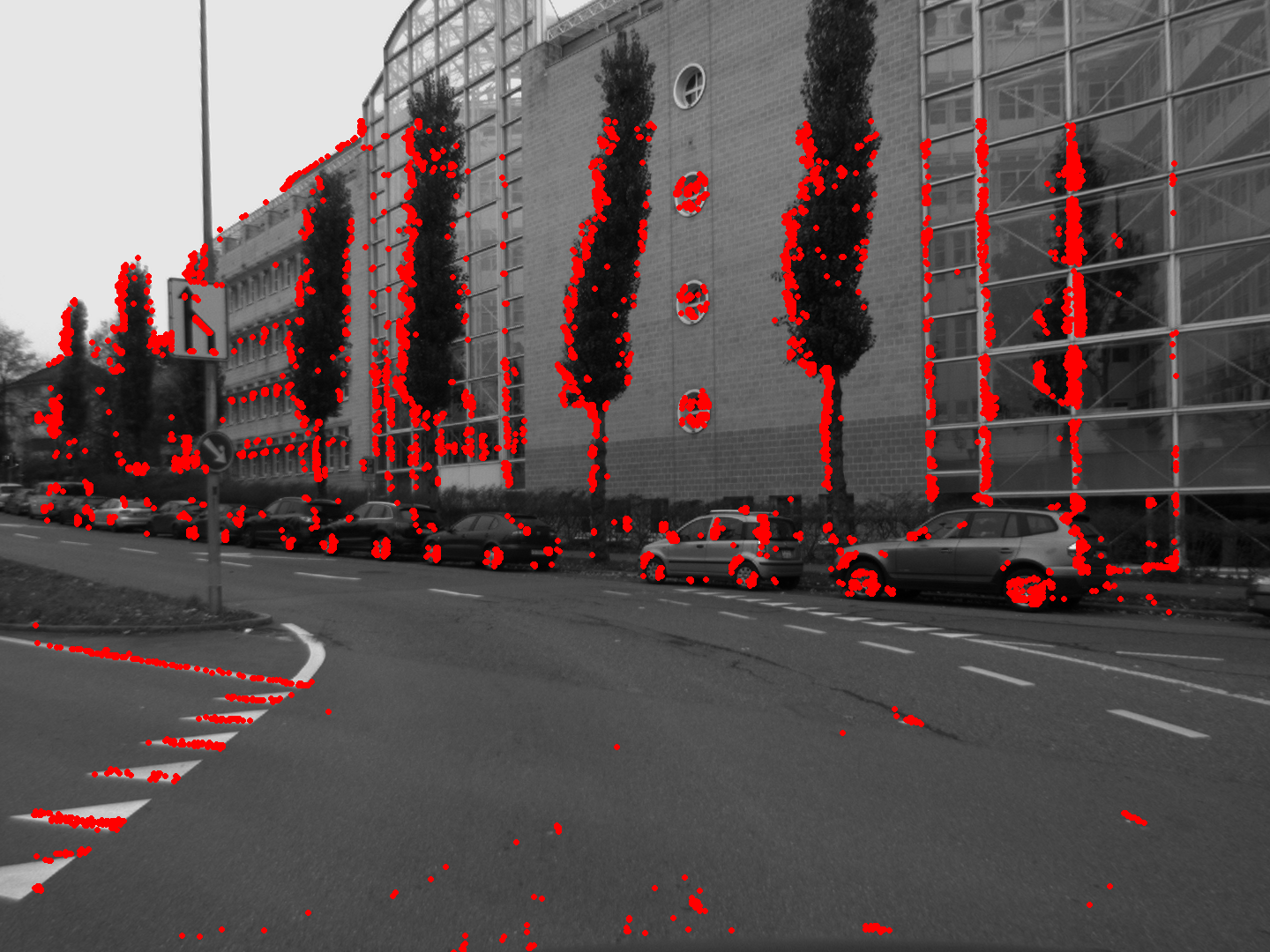}}
 \caption{
 \textit{Point sampling results of ESVO and our method.}
 (a) and (b) are ESVO's results that have 10k points and 5k points sampled, respectively.
 (c) is our sampling result.}
    \label{fig:sampling result comparison}
\vspace{-4.5em}
\end{figure}

The pixel value in an AA map represents the number of events that have been accumulated at this pixel over the locally, adaptively controlled time interval.
These pixels with higher values are more likely to be associated to instantaneous edge pixels.
To evaluate the effectiveness of the proposed AA method, we compare its result against another three image-like representations, including time surfaces (TS)~\cite{Delbruck08issle} and two speed-invariant representations~\cite{Manderscheid19cvpr, glover2021luvharris}.
As illustrated in Fig.~\ref{fig:image-like representation comparision}, the result of AA preserves relatively complete edges with the least redundant points while keeping the highest signal-to-noise ratio.
Edge pixels used by the following mapping operation are sampled from each patch independently. 
In general, the bigger the AA pixel value in each patch, the higher possibility the pixel is selected.
We further shuffle the sampling pixels to assure an even sampling.
We compare the sampling result with edge pixels obtained from ESVO.
As shown in Fig.~\ref{fig:sampling result comparison}, our sampled pixels are more evenly and continuously distributed on the edge structures.

\begin{algorithm}[t]
\caption{Adaptive Accumulation of Events }
\label{alg:AA algorithm}
\renewcommand{\algorithmicrequire}{\textbf{Input:}}
\renewcommand{\algorithmicensure}{\textbf{Output:}}
\begin{algorithmic}[1]
\REQUIRE
$N_e$ Events $\{e_k \doteq (x_k, y_k, t_k, p_k)\}_{k=1}^{N_e}$, parameters $\beta$.\\
\ENSURE
Adaptive accumulation map $\mathbf{A}(x, y)$.
\STATE Initialize $\mathbf{A}(x, y)$ with all zero elements, and  $t_\text{last} = 0$. 
\STATE Divide $\mathbf{A}(x, y)$ to N blocks $\{\mathbf{A}_{i} ~|~i = 1, 2, \cdots N\}$, and assign each block with a boolean flag $\mathcal{F}_i$. 
\STATE Set $\{\mathcal{F}_i = True~|~ i = 1, 2, \cdots N\} $.
\FOR{$k=1,\cdots,N_e$}
\STATE Get block $\mathbf{A}_{i}$ according to the coordinate of $e_{k}$.
\IF{ $\mathcal{F}_{i}$ != True  }
\STATE continue.
\ENDIF
\STATE $\mathbf{A}_{i}(x_k, y_k)$++.
\IF{\textbf{ImageContrast}($\mathbf{A}_{i}$) $ > \beta$}
\STATE $\mathcal{F}_{i}$ = False.
\ENDIF
\ENDFOR
\end{algorithmic}
\end{algorithm}

\subsection{Mapping}
\label{subsec:mapping}

The static stereo method in \cite{zhou2021event} can hardly recover accurate depth of structures that are parallel to the baseline of the stereo camera.
This is because the spatio-temporal profile of these structures are only distinctive in one direction, and thus, many false-positive matches will be witnessed during the epipolar-line searching.
On the contrary, temporal stereo methods are unlikely affected by this issue.
As long as the stereo camera does not move along the baseline, epipolar lines defined between a temporal stereo pair are no longer parallel to the baseline of the static stereo.
This assumption always holds for forward looking stereo cameras, \eg,~those used in driving scenes.
Inspired by \cite{Engel15iros}, we introduce the temporal stereo method to solve this problem.

Let's consider the normally applied horizontal stereo configuration.
We divide the sampled edge pixels from Sec.~\ref{subsec:Adaptive Accumulation of Events} into two groups according to their gradient direction on the corresponding TS.
In the first group, we collect pixels at which the magnitude ratio ($\eta$) between the horizontal gradient and the vertical gradient is smaller than a threshold.
The remaining sampled pixels make up the second group, and we feed them to the mapping method of the original ESVO.
The first group is fed to the proposed temporal stereo method as discussed in the following.

The key to the event-based temporal stereo is effectively exploiting appearance similarity in the spatio-temporal profile.
This requires the event representation, on which the stereo data association is established, to possess the speed-invariant property.
Thus, time surfaces used in ESVO are no longer applicable in this context.
We investigate the temporal stereo matching performance on three representations, including TOS~\cite{glover2021luvharris}, SILC~\cite{Manderscheid19cvpr} and our AA.
We find AA is the optimal choice because of its higher signal-noise ratio.

\begin{figure}[t]
	\centering
	\subfigure[Temporal stereo on AA.]{\includegraphics[width=0.23\textwidth]{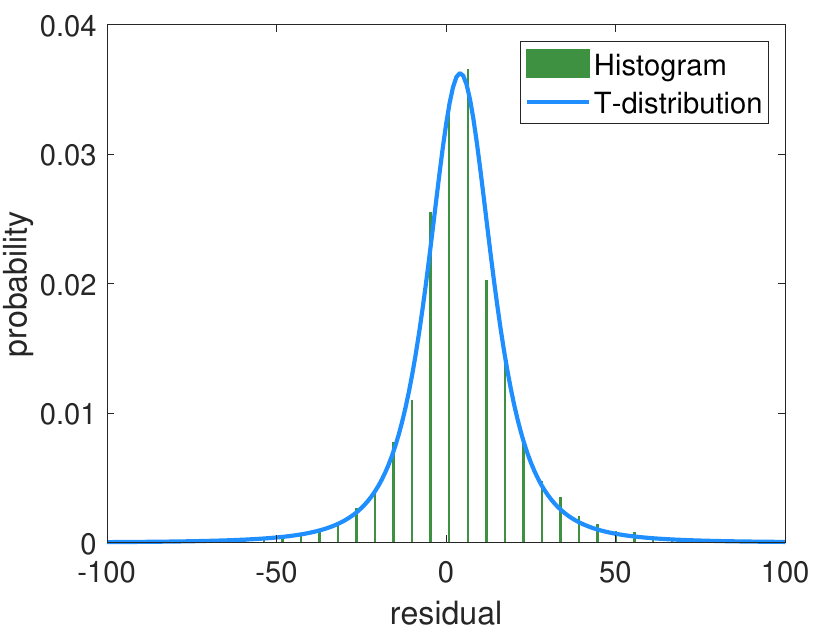}}\hspace{1pt}
    \subfigure[Static stereo on TS.]{\includegraphics[width=0.23\textwidth]{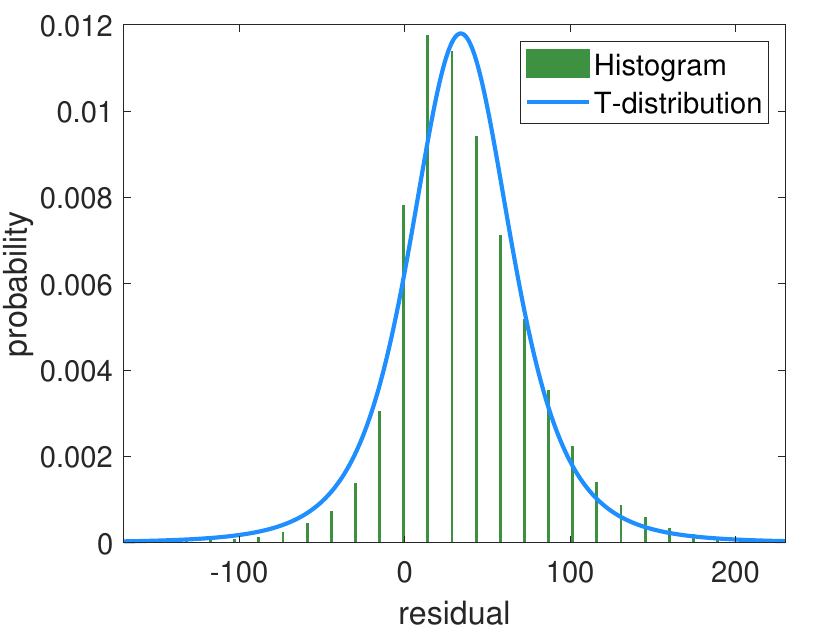}}
	\caption{
 \textit{Probability distribution (PDF) of the temporal stereo residuals $r_\text{temporal}$ and static stereo residuals $r_\text{static}$.} 
 The curves (blue) are the $\textit{Student’s t}$ fitting results from the empirical histogram (green).
 }
    \label{fig:probabilistic characteristics of temporal stereo}
\vspace{-1.em}
\end{figure}

Given as prior the relative pose between a temporal stereo pair, the proposed temporal stereo method imitates ESVO's mapping strategy in the sense of applying a block matching plus a nonlinear refinement.
To fuse multiple temporal stereo estimates, we follow the way in \cite{zhou2021event} to obtain the probabilistic characteristics of the temporal stereo results.
As shown in Fig.~\ref{fig:probabilistic characteristics of temporal stereo}, 
both the temporal stereo residuals $r_{\text{temporal}}$ evaluated on AAs and the static stereo residuals $r_{\text{static}}$ evaluated on TSs approximately obey the \textit{student's t} distribution
\begin{equation}
\label{eq:t distribution}
    r \sim St(\mu_{r}, s_{r}, \nu_{r}),
\end{equation}
where $\mu_{r}$, $s_{r}$, $\nu_{r}$ are the model parameters, namely the mean, scale and degree of freedom.
Although the depth estimates from the two methods are probabilistically compatible, they cannot be fused straightforwardly.
This is due to the uncertainty of the temporal stereo's result is always much smaller, and it is just caused by the different nature of the heterogeneous representations.
To obtain a more complete depth map, we simply merge the results of the two stereo methods.
We show the mapping results in Sec.~\ref{subsec:comparison of mapping}.

\subsection{Tracking}
\label{subsec:tracking}

Our tracking module basically follows the ESVO framework, which takes as input the events, a TS and a local 3D map, and computes the pose of the stereo rig with respect to the map.
Let $\mathcal{S}^{\mathcal{F}_{\mathbf{ref}}} = \{ \mathbf{x}_{i}\}$ represent a set of pixels with inverse depth values in the reference frame, and $\mathcal{T}(\mathbf{x}, t)$ and $\overline{\mathcal{T}}(\mathbf{x}, t) = 1 - \mathcal{T}(\mathbf{x}, t)$ be the TS and negative TS at time $t$.
The purpose of tracking is, identical to \cite{zhou2021event}, to determine the optimal motion parameters $\boldsymbol{\theta}$ by solving
\begin{equation}
\label{eq: optimization}
\boldsymbol{\theta}^{\ast} = \mathop{\arg\min}\limits_{\boldsymbol{\theta}} \sum\limits_{\mathbf{x} \in \mathcal{S}^{\mathcal{F}_{\mathbf{ref}}}} \overline{\mathcal{T}}_{\mathbf{left}}(\boldsymbol{W}(\mathbf{x}, \rho; \boldsymbol{\theta})),
\end{equation}
where the warp function $ \boldsymbol{W}(\mathbf{x}, \rho; \boldsymbol{\theta})$ denotes the transformation from local depth points to the latest negative TS image.
And $\boldsymbol{\theta} \doteq (\mathbf{c}^{\mathsf{T}}, \mathbf{t}^{\mathsf{T}})^{\mathsf{T}}$ are the motion parameters, where $\bc = (c_{1},c_{2},c_{3})^{\mathsf{T}} $ are the Cayley parameters ~\cite{cayleyparameter} for rotation and $\mathbf{t} = (t_{x}, t_{y},t_{z})^{\mathsf{T}} $ is the translation.

To improve the accuracy of pose estimation, we leverage the pre-integration of gyroscope measurements to provide an initial value of rotation in Eq.~\ref{eq: optimization}.
For an IMU with a 3-axis gyroscope, the measurement of IMU angular velocity can be represented by
\begin{equation}
\label{eq:gyro measurement}
  \tilde{\boldsymbol{\omega}}^{b} = \boldsymbol{\omega}^{b} + \mathbf{b}_{g}^{b} +  \mathbf{n}_{g}^{b},  
\end{equation}
where $\mathbf{b}_{g}^{b}$, $\mathbf{n}_{g}^{b}$ are the bias and noise of the gyroscope expressed in the IMU frame.
The relative rotation between two successive tracking estimates, \eg,~from time $t_i$ to $t_{i+1}$, expressed in the IMU frame $b_{i}$, can be calculated by
\begin{equation}
\label{eq:integration of angular velocity}
    \boldsymbol{\gamma}_{b_{i+1}}^{b_{i}} = \int_{t \in [t_{i}, t_{i+1}]} \frac{1}{2} \boldsymbol{\Omega}(\tilde{\boldsymbol{\omega}}^{b_{t}} - \mathbf{b}_{g}^{b_{t}} - \mathbf{n}_{g}^{b_{t}}) \boldsymbol{\gamma}_{b_{t}}^{b_{i}} \textit{dt},
\end{equation}
where $\boldsymbol{\gamma}_{b_{t}}^{b_{i}}$ is the quaternion representation of the relative rotation, and $\boldsymbol{\Omega}(\boldsymbol{\omega}) = \begin{bmatrix}  - {\lfloor \boldsymbol{\omega} \rfloor}_{\times} & \boldsymbol{\omega} \\  -\boldsymbol{\omega}^{\mathsf{T}} & 0 \end{bmatrix}$.
The bias $\mathbf{b}_{g}^{b}$ is initialized empirically and not updated throughout this work.
The initial rotation parameter $\bc_0$ is obtained via a quaternion-Cayley transformation.
We show the benefit brought by using the pre-integration result as a motion prior in Sec.~\ref{subsec:system evaluation}.

\subsection{System}
\label{subsec:system}
\begin{figure}[t]
    \centering
    \includegraphics[width=0.48\textwidth]{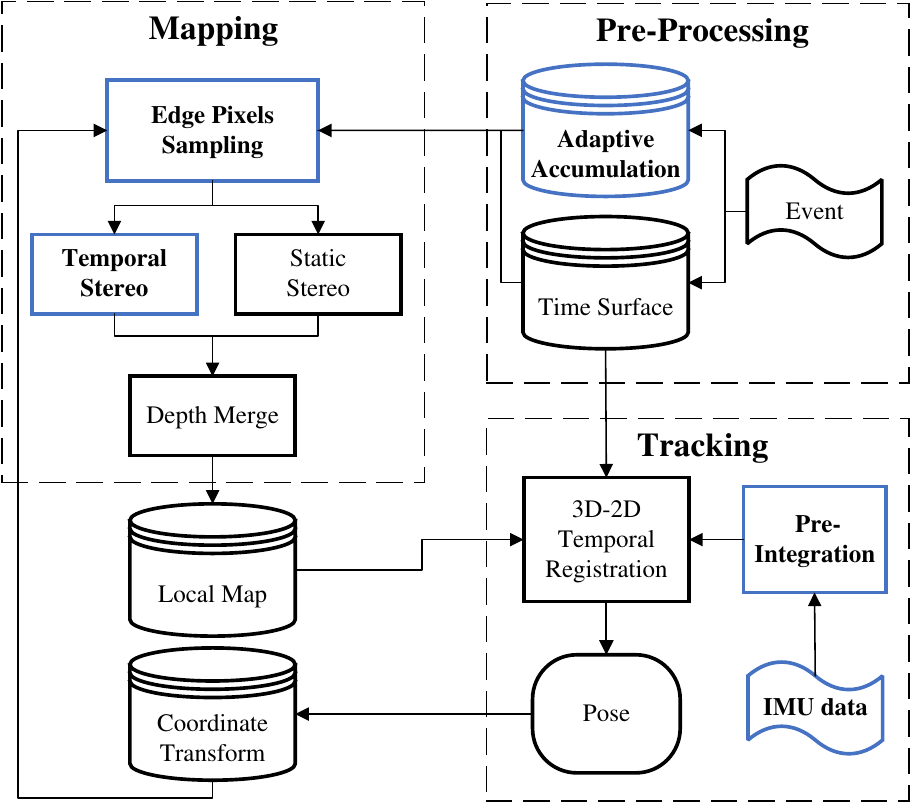}
    \caption{\textit{Flowchart of the proposed system}. 
    New functions added to the original \textit{ESVO} framework are highlighted in blue.
    Each module enclosed by a dashed box is executed independently and occupies at least one thread.}
    \label{fig:system flowchart}
    \vspace{-1.em}
\end{figure}

We extend ESVO~\cite{zhou2021event} with several additional modules.
As shown in Fig.~\ref{fig:system flowchart}, the whole system takes as input the events from a stereo event camera and gyroscope measurements from an IMU.
In the pre-processing module, the AA and TS are generated at a fixed rate (e.g., 100 Hz), and they are fed to the mapping module together with the tracking results.
In the mapping module, the sampled edge pixels are divided according to the classification parameter $\eta$, and the results of the temporal stereo method and static stereo method are merged and inserted to the local map.
Given the local map, the most recent TS, and also the initial rotation guess from IMU pre-integration, the tracking module calculates the camera pose.
All hyper parameters used are set as in Table.~\ref{tab:parameter value setting}.

\begin{table}[t]
    \centering
    \caption{Settings of hyper parameters.}
    \setlength{\tabcolsep}{3mm}{
    \begin{tabular}{ccccc}
    \hline
        Parameter & $w$ & $\delta t$ & $\beta$ & $\eta$ \\
    \hline
        \multirow{2}{*}{Value} & 80 pixel (for \textit{DSEC}) & \multirow{2}{*}{2 ms} & \multirow{2}{*}{0.5} & \multirow{2}{*}{0.2}  \\
        {~} & 30 pixel (for \textit{rpg}) & {~} & {~} & {~} \\ 
    \hline
    \end{tabular}}
    \label{tab:parameter value setting}
    \vspace{0.5em}
\end{table}

\section{Experiments}
\label{sec:evaluation}

We evaluate our method in this section using two publicly available datasets, collected using a hand-held stereo event camera (\textit{rpg} dataset)~\cite{Zhou18eccv} and a mobile-mounted stereo event camera (\textit{DSEC} dataset)~\cite{Gehrig21ral}, respectively.
In this section, we first evaluate our solution to the mapping sub-problem quantitatively and qualitatively (Sec.~\ref{subsec:comparison of mapping}).
Second, we test the full system by evaluating the recovered trajectories (Sec.~\ref{subsec:system evaluation}).
Finally, we present an analysis on the computational efficiency (Sec.~\ref{subsec:implementation details and computational efficiency}).

\subsection{Comparison of Mapping: ESVO vs Ours}
\label{subsec:comparison of mapping}
\global\long\def\figWidth{0.13\linewidth} 
\begin{figure}[t]
\centering
\setlength\tabcolsep{1pt}
\begin{tabular}{ccccc}
    {} & Intensity image & ESVO & Our method  \\ 
    \rotatebox{90}{\makecell{dsec\_city04\_a}}&
    \begin{minipage}[b]{0.3\columnwidth}
		\centering
		{\includegraphics[width=\linewidth]{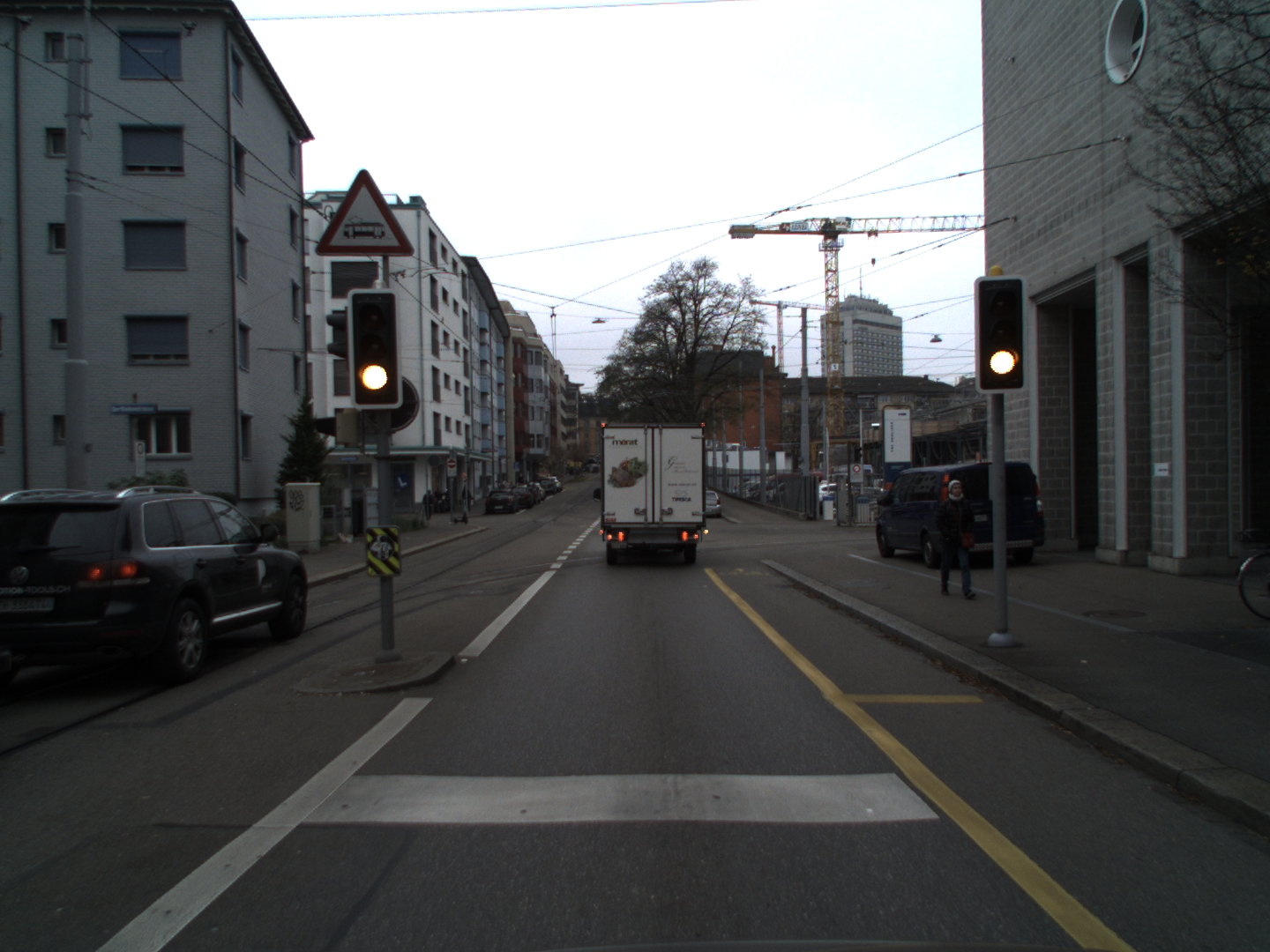}}
    \end{minipage} &     
	\begin{minipage}[b]{0.3\columnwidth}
		\centering
		{\includegraphics[width=\linewidth]{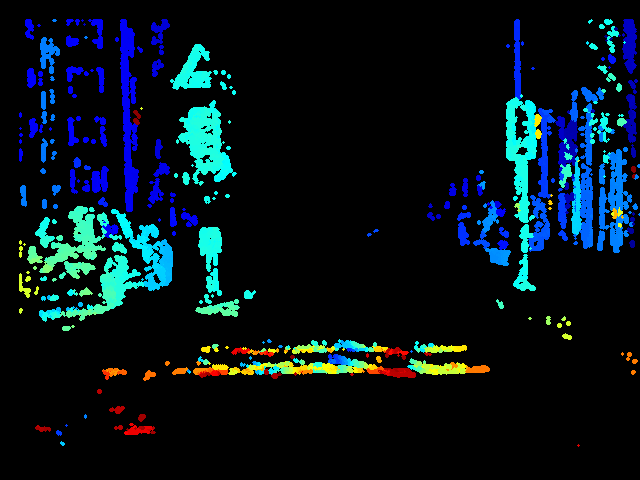}}
	\end{minipage} &      
	\begin{minipage}[b]{0.3\columnwidth}
		\centering
		{\includegraphics[width=\linewidth]{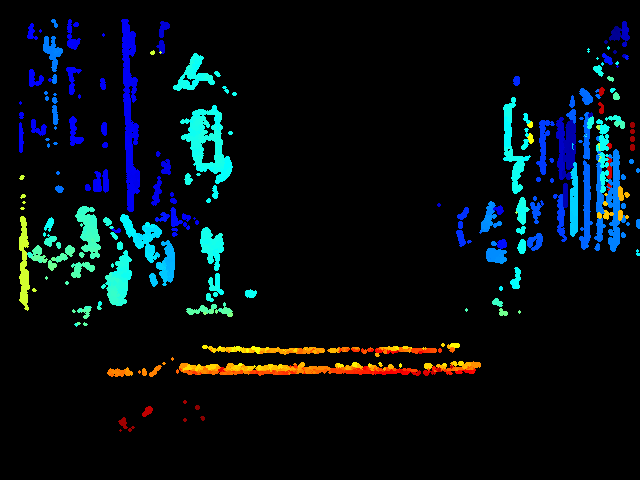}}
	\end{minipage}  \vspace{2pt}\\
    \rotatebox{90}{\makecell{dsec\_city04\_c}}&
    \begin{minipage}[b]{0.3\columnwidth}
		\centering
		{\includegraphics[width=\linewidth]{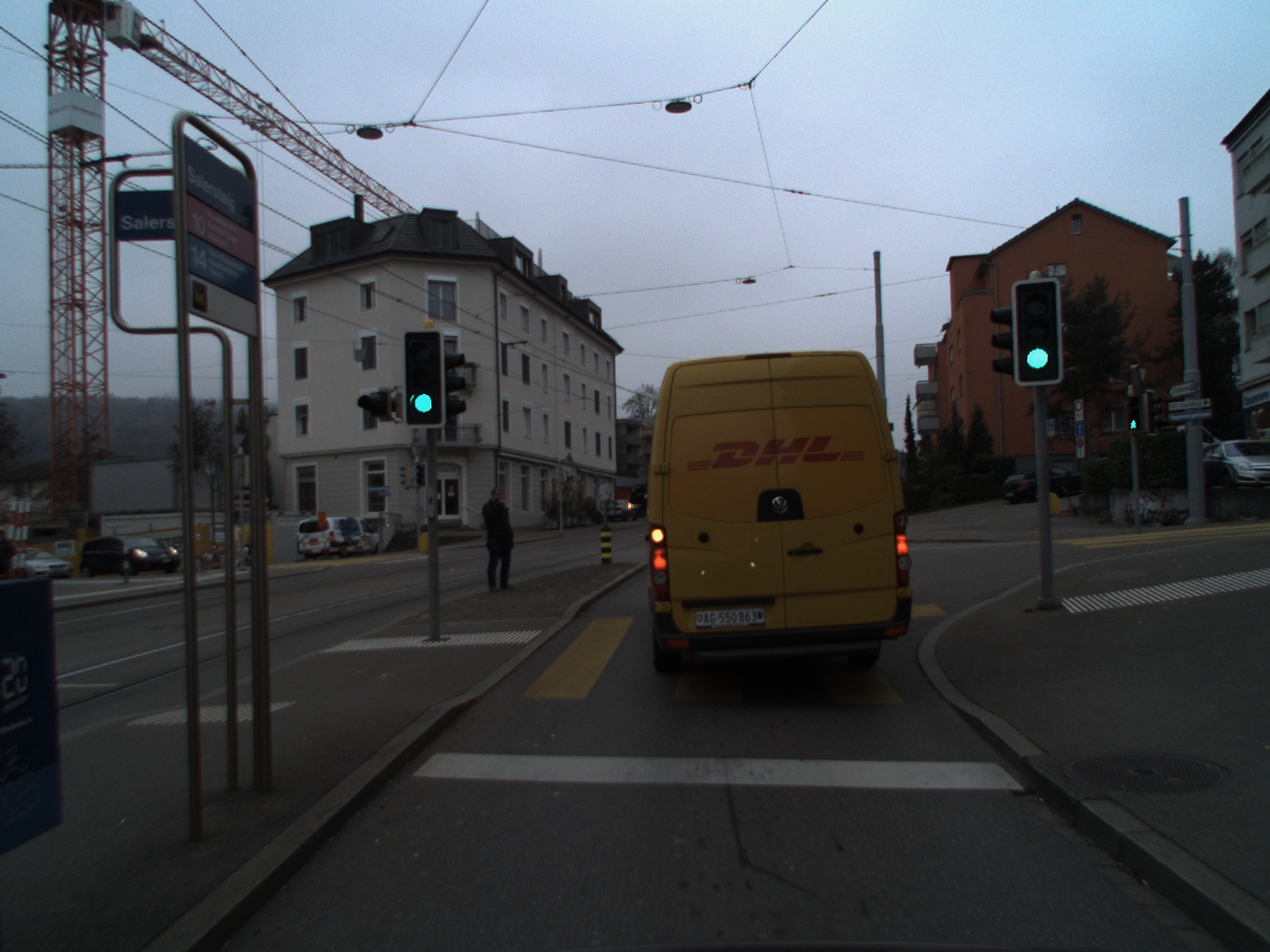}}
    \end{minipage} &     
	\begin{minipage}[b]{0.3\columnwidth}
		\centering
		{\includegraphics[width=\linewidth]{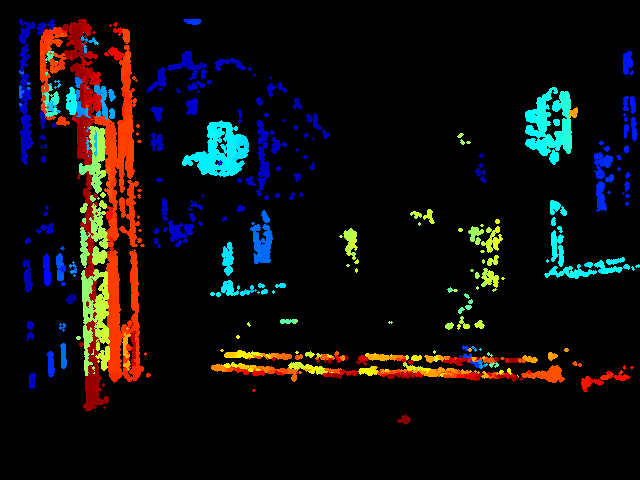}}
	\end{minipage} &      
	\begin{minipage}[b]{0.3\columnwidth}
		\centering
		{\includegraphics[width=\linewidth]{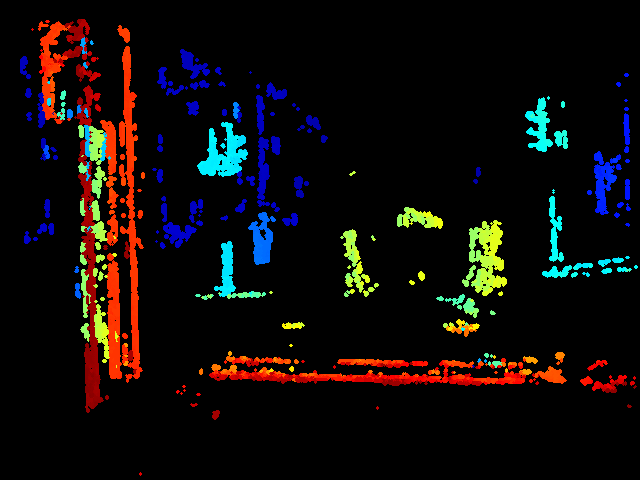}}
	\end{minipage}  \vspace{2pt}\\
    \rotatebox{90}{\makecell{rpg\_reader}}&
    \begin{minipage}[b]{0.3\columnwidth}
		\centering
		{\includegraphics[width=\linewidth]{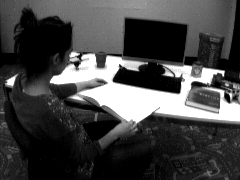}}
    \end{minipage} &     
	\begin{minipage}[b]{0.3\columnwidth}
		\centering
		{\includegraphics[width=\linewidth]{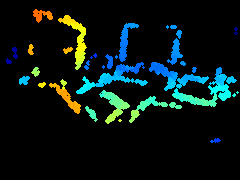}}
	\end{minipage} &      
	\begin{minipage}[b]{0.3\columnwidth}
		\centering
		{\includegraphics[width=\linewidth]{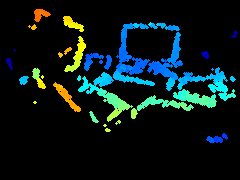}}
	\end{minipage}  \vspace{2pt}\\
\end{tabular}
\caption{\textit{Qualitative comparison of mapping results}.
Intensity images in the first column are used only for visualization.
The second and third columns show the estimated inverse depth map by ESVO and our method, respectively. 
Note that our method returns more accurate and complete reconstruction results for the horizontal edges.
Depth maps are color coded, from red (close) to blue (far) over a black background, in the range 1 m-50 m for the \textit{DSEC} dataset and 0.5 m-10 m for the \textit{rpg} dataset.
}
\label{fig:mapping evaluation on DSEC/RPG}
\vspace{-2.5em}
\end{figure}
\begin{table}[t]
\vspace{-0em}
\begin{center}
\caption{\textit{Quantitative comparison of mapping result.}
The depth range refers to the average of true depth from points evaluated.}
\vspace{0em}
\setlength{\tabcolsep}{3mm}{
\begin{tabular}{cccc}
\hline
{Sequence~(depth range)} & {} & {ESVO} & {Ours}\\
\hline
\multirow{3}{*}{\makecell[c]{dsec\_city04\_a \\ (9.95 m)}} & Mean error & 0.66 m & \textbf{0.41 m}\\
{~}& Median error & 0.43 m & \textbf{0.32 m}\\
{~}& Relative error & 7.8\% & \textbf{4.3\%} \\
\specialrule{0em}{1pt}{1pt}
\multirow{3}{*}{\makecell[c]{dsec\_city04\_c \\ (6.86 m)}} & Mean error & 0.83 m & \textbf{0.69 m}\\
{~} & Median error & 0.33 m & \textbf{0.28 m}\\
{~} & Relative error & 15.3\% & \textbf{11.6\%} \\ 
\specialrule{0em}{1pt}{1pt}
\multirow{3}{*}{\makecell[c]{dsec\_city04\_d \\ (14.61 m)}} & Mean error & 1.01 m & \textbf{0.65 m}\\
{~} & Median error & 0.65 m & \textbf{0.58 m}\\
{~} & Relative error & 11.2\% & \textbf{7.1\%} \\ 
\hline
\end{tabular}
}
\label{tab:mapping evaluation}
\end{center}
\end{table}

We compare our mapping results against those of the original ESVO pipeline \cite{zhou2021event}.
As shown in Fig.~\ref{fig:mapping evaluation on DSEC/RPG}, our results achieve better performance in terms of reconstruction completeness and local depth smoothness.
In the results of ESVO, we observe that horizontal structures are typically neither recovered nor estimated accurately.
This is due to the aperture problem encountered by the original static stereo matching operation.
Compared to the original ESVO pipeline \cite{zhou2021event}, our solution to the mapping sub-problem introduces the temporal stereo estimation between successive observations of the left event camera.
This additional stereo matching operation recovers 3D information missed (or inaccurately estimated) by the static stereo operation.
We also carry out a quantitative evaluation on the mapping results and use the relative depth error \cite{Zhou18eccv} as the evaluation metric.
As shown in Table.~\ref{tab:mapping evaluation}, our method outperforms ESVO in terms of mean error, median error, and average relative error in depth estimation.

\subsection{Full System Evaluation: ESVO vs Ours}
\label{subsec:system evaluation}
We evaluate the full system using two publicly available datasets.
The first one is the \textit{rpg} dataset, which features a hand-held stereo event camera moving in a small-scale indoor environment.
We report ego-motion estimation results using two standard metrics: relative pose error (RPE) and absolute trajectory error (ATE) \cite{Sturm12iros}, and the results are given as root-mean-square errors (RMSEs). 
As shown in Table.~\ref{tab:dsec_rpe_eval} and Table.~\ref{tab:dsec_ate_eval}, our new pipeline outperforms ESVO in terms of these two metrics.

\begin{table}[t]
\vspace{-0em}
\begin{center}
\caption{Translation [\%] and rotation [°/m] evaluation results of the proposed method compared to ESVO using relative pose RMSE.}
\vspace{0em}
\setlength{\tabcolsep}{2.5mm}{
\begin{tabular}{ccccc}
\hline
\multirow{2}{*}{Sequence} &\multicolumn{2}{c}{ESVO} &\multicolumn{2}{c}{Ours}\\
\cline{2-5}
{~}&$\mathbf{R}$&$\mathbf{t}$&$\mathbf{R}$&$\mathbf{t}$\\
\hline
rpg\_box & 1.92 & 3.79 & \textbf{0.82} & \textbf{1.88}\\
rpg\_monitor & 3.30 & 5.62 & \textbf{1.93} & \textbf{3.05}\\
rpg\_reader & \textbf{3.32} & 11.98 & 3.68 & \textbf{8.46}\\
\hline 
dsec\_city04\_a & 0.10 & 8.12 & \textbf{0.08} & \textbf{3.46} \\
dsec\_city04\_b & \textbf{0.13} & 8.46 & 0.21 & \textbf{6.53} \\
dsec\_city04\_c & \textbf{0.04} & 8.75 & 0.05 & \textbf{6.47} \\
dsec\_city04\_d & 0.08 & 16.17  & \textbf{0.05} & \textbf{4.48} \\
dsec\_city04\_e & 0.16 & 21.17 & \textbf{0.08} & \textbf{5.57} \\
dsec\_city11\_a & \textbf{0.06} & 9.79 & 0.07 & \textbf{2.96} \\
\hline
\end{tabular}
}
\label{tab:dsec_rpe_eval}
\end{center}
\vspace{-4em}
\end{table}

\begin{table}[t]
\vspace{-0em}
\begin{center}
\caption{Absolute trajectory RMSE [$\mathbf{t}$:cm]}
\vspace{0em}
\setlength{\tabcolsep}{4mm}{
\begin{tabular}{cccc}
\hline
{Sequence} & {ESVO} & {Ours}\\
\hline
rpg\_box & 9.5 & \textbf{5.0} \\
rpg\_monitor & 5.8 & \textbf{2.8}\\
rpg\_reader & 6.6 & \textbf{3.7} \\
\hline
dsec\_city04\_a & 371.1 & \textbf{105.0} \\
dsec\_city04\_b & 116.6 & \textbf{66.7} \\
dsec\_city04\_c & 1357.1 & \textbf{637.9} \\
dsec\_city04\_d & 2676.6 & \textbf{699.8} \\
dsec\_city04\_e & 794.9 & \textbf{130.3} \\
dsec\_city11\_a & 364.0 & \textbf{92.7} \\
\hline
\end{tabular}
}
\label{tab:dsec_ate_eval}
\end{center}
\end{table}

\begin{figure}[t]
	\centering
	\subfigure[dsec\_city\_04\_a]{\includegraphics[width=0.23\textwidth]{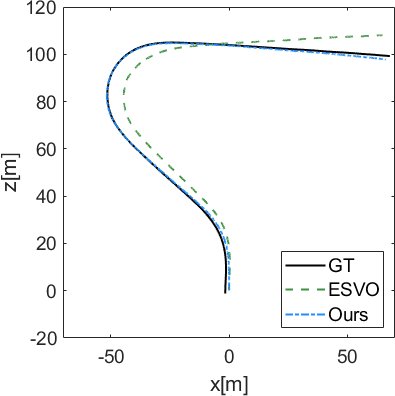}}\hspace{0pt}
	\subfigure[dsec\_city\_04\_b]{\includegraphics[width=0.23\textwidth]{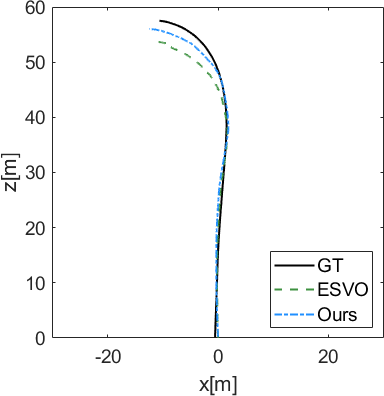}}\\
        \subfigure[dsec\_city\_04\_c]{\includegraphics[width=0.23\textwidth]{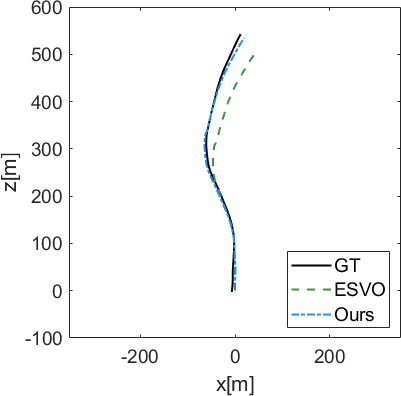}}\hspace{0pt}
	\subfigure[dsec\_city\_04\_d]{\includegraphics[width=0.23\textwidth]{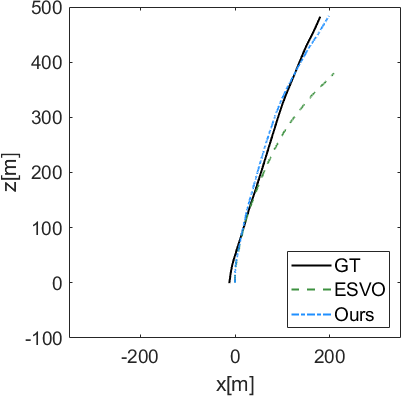}}\\
        \subfigure[dsec\_city\_04\_e]{\includegraphics[width=0.23\textwidth]{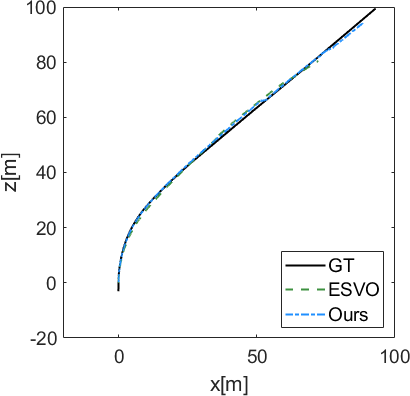}}\hspace{0pt}
	\subfigure[dsec\_city\_11\_a]{\includegraphics[width=0.23\textwidth]{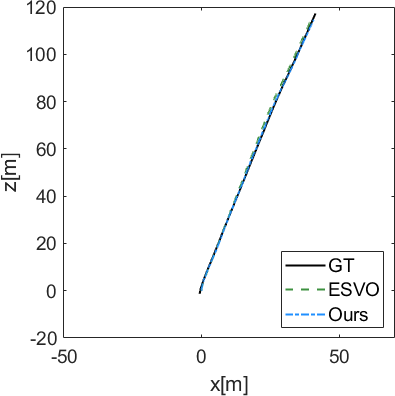}}\\
	\caption{Illustration of recovered trajectories.}
    \label{fig:traj_eval}
\end{figure}

The second dataset used is \textit{DSEC}, which is a stereo event camera dataset for large-scale driving scenarios.
We first demonstrate the benefit brought by using inertial measurements as a prior in the camera pose tracking sub-problem.
As discussed in Sec.~\ref{subsec:tracking}, the introduction of inertial measurements alleviates the problem of 3D-2D spatio-temporal registration being insensitive to recovering rotation in general 6-DoF motion.
This improvement is clearly witnessed when the stereo event camera undergoes a translation plus a rotation in the yaw axis.
This is justified by the relative pose RMSE results shown in Table.~\ref{tab:imu benefit},
where two configurations (with and without IMU) are compared.
Additionally, we compare extensively our results against that of ESVO pipeline on the \textit{DSEC} dataset.
We apply the evaluation tool provided in \cite{zhang2018tutorial} and also report the absolute trajectory error and relative pose error.
As shown in Table.~\ref{tab:dsec_rpe_eval} and Table.~\ref{tab:dsec_ate_eval}, our new pipeline outperforms ESVO in terms of both ATE and RPE.
We also illustrate the resulting trajectories against the groundtruth (GT).
As shown in Fig.~\ref{fig:traj_eval}, our trajectories (Ours) are typically more consistent with the GT compared to those of ESVO.

\begin{table}[t]
\vspace{-0em}
\begin{center}
\caption{Translation [\%] and rotation [°/m] evaluation results of the proposed method compared to w/o IMU using relative pose RMSE.}
\vspace{0em}
\setlength{\tabcolsep}{2.5mm}{
\begin{tabular}{ccccccc}
\hline
\multirow{2}{*}{Sequence} &\multicolumn{2}{c}{w/o IMU} &\multicolumn{2}{c}{w/ IMU}\\
\cline{2-5}
{~}&$\mathbf{R}$&$\mathbf{t}$&$\mathbf{R}$&$\mathbf{t}$\\
\hline
dsec\_city04\_a & 0.13 & 4.06 & \textbf{0.11} & \textbf{3.84} \\
dsec\_city04\_c & 0.06 & 7.71 & \textbf{0.05} & \textbf{6.47} \\
\hline
\end{tabular}
}
\label{tab:imu benefit}
\end{center}
\end{table}

\begin{table}[!t]
\vspace{-0em}
\begin{center}
\renewcommand\arraystretch{1.5}
\caption{Computational performance [time:ms]}
\vspace{0em}
\setlength{\tabcolsep}{0.5mm}{
\begin{tabular}{cccc}
\hline
{Node (\#Threads)} & {Function} & {ESVO} & {Ours}\\
\hline
\multirow{2}{*}{Pre-processing (1)} & TS & 27 ($\sim$70k) & \textbf{3 ($\sim$70k)}\\
{~} & AA & - & 8 ($\sim$70k) \\

\hline
\multirow{1}{*}{Tracking (2)} & Non-linear solver & 8 ($\sim$2k) & \textbf{7 ($\sim$2k)} \\

\hline
\multirow{5}{*}{Mapping (4)} & Event matching & 36 ($\sim$10k) & \textbf{18 ($\sim$2.5k)} \\
{~} & Depth optimization & 82 ($\sim$4.5k) & \textbf{8 ($\sim$0.9k)} \\
{~} & Depth fusion & 23 ($\sim$140k) & \textbf{4 ($\sim$12k)} \\
\cdashline{2-4}
{~} & Regularization (optional) & 296 ($\sim$25k) & - \\
{~} & {Total (w/ optional)} & 141 (437) & \textbf{30}  \\
\hline
\end{tabular}
}
\label{tab:6}
\vspace{-5em}
\end{center}
\end{table}

\subsection{Computational Efficiency}
\label{subsec:implementation details and computational efficiency}

As shown in Table.~\ref{tab:6}, We compare the computational performance between ESVO and our method using a desktop with Intel Core i7-12700K on the \textit{DSEC}. 
The reported runtime for each functionality is evaluated on a rough number of points, denoted by ($\sim$).
Both systems are accelerated using hyper-threading technology, and the number of threads occupied by each node are declared inside the parentheses.
We improve the generation method of TS by reorganizing data structure, making it 9 times faster than that in ESVO.
The efficiency of tracking is also slightly improved due to initializing the pose parameters with relative rotation from gyroscope pre-integration.
Note that the number of points used for mapping is different between two systems. 
The minimum number of points required by ESVO's mapping for a normal performance is 10k, and our mapping method requires only 2.5k.
In this way, our mapping takes almost one fifth of the time of ESVO and still has remarkably better tracking performance. 
Additionally, regularization is very time-consuming and has very little impact on tracking performance, and thus, it is used optionally.

\section{Conclusion}
\label{sec:conclusion}
We present an IMU-aided event-based stereo visual odometry system in this work.
Built on top of ESVO, a state-of-the-art event-based visual odometry pipeline, our framework additionally introduces three modules, namely the efficient edge-pixel sampling strategy, the temporal stereo mapping operation, and the usage of inertial measurements.
The first two modules lead to more complete and accurate mapping results, and the third module improves the accuracy of camera pose tracking.
Besides, our framework scales better with event streaming rate featured by modern event cameras with a high spatial resolution (\eg, $640 \times 480$ pixel), indicating a step forward of direct methods for practical applications.

\bibliographystyle{IEEEtran}
\bibliography{myBib}
\end{document}